\documentclass[10pt]{article}
\usepackage[preprint]{tmlr}

\usepackage{amsmath,amsfonts,bm}

\def\eqref#1{equation~\ref{#1}}

\def\1{\bm{1}}

\DeclareMathAlphabet{\mathsfit}{\encodingdefault}{\sfdefault}{m}{sl}
\SetMathAlphabet{\mathsfit}{bold}{\encodingdefault}{\sfdefault}{bx}{n}

\usepackage{hyperref}
\usepackage{url}
\usepackage{booktabs}
\usepackage{multirow}
\usepackage{array}
\usepackage{tabularx}
\usepackage{graphicx}
\usepackage{subcaption}
\usepackage{placeins}
\usepackage{xcolor}

\graphicspath{{figures/imagegen/}}

\title{Numeracy in Large Language Models: \\Fundamental Limitations and Paths to Improvement}

\author{\name Aoxin Ni \email aoxin.ni@ucas.ac.cn \\
      \addr Department of Computer Science\\
      University of Chinese Academy of Sciences}

\hypersetup{
  hidelinks,
  pdftitle={Numeracy in Large Language Models: Fundamental Limitations and Paths to Improvement},
  pdfauthor={Aoxin Ni},
  pdfsubject={Numeracy in large language models},
  pdfkeywords={large language models, numeracy, numerical grounding, arithmetic reasoning}
}

\def\month{MM}
\def\year{YYYY}
\def\openreview{\url{https://openreview.net/forum?id=XXXX}}

\begin{document}

\maketitle

\begin{abstract}
Despite remarkable progress in mathematical reasoning---achieving near-human performance on
competitive benchmarks such as MATH, GSM8K, and Olympiad-level problems---large language
models (LLMs) continue to exhibit astonishing failures on the most elementary numerical tasks.
Simple magnitude comparisons, addition of large integers, basic fraction arithmetic, and
operations in scientific notation routinely produce incorrect results across frontier models.
This survey provides a focused and comprehensive overview of the emerging field of basic
numerical understanding in large language models, explicitly distinguishing it from the broader
topic of mathematical reasoning. We introduce the \emph{Numerical Grounding Framework} (NGF),
an original two-part theoretical construct that decomposes numeracy into \emph{Representational
Grounding} (RG)---the faithful mapping of numeral surface forms to value, magnitude, and
format-equivalent internal representations---and \emph{Procedural Grounding} (PG)---the
faithful execution of arithmetic procedures consistent with their mathematical definitions.
Building on Harnad's (1990) symbol grounding problem and Dehaene's (2011) cognitive theory of
number sense, NGF provides a unifying lens that organizes failure modes, root causes, and
mitigation strategies under a single conceptual umbrella. We survey the new generation of
specialized diagnostic benchmarks that emerged in 2024--2025, analyze structural root causes in
BPE tokenization, positional encoding, embedding geometry, pretraining data distribution, and
critically evaluate proposed mitigation strategies. A key finding replicated across multiple
studies is that architectural interventions effective for models trained from scratch---such as
Little-Endian fine-tuning and Abacus Embeddings---are frequently inapplicable to already-pretrained
large models, for which supervised fine-tuning on diverse numerical examples and inference-time
scaffolding remain the most consistently effective approaches. We conclude with practical
deployment recommendations and an agenda for achieving robust, human-like number sense in future
foundation models.
\end{abstract}

\section{Introduction}

The evolution of Large Language Models (LLMs) has transcended the boundaries of natural
language generation, catalyzing a paradigm shift across specialized domains that demand
rigorous quantitative reasoning. In the realm of mathematical reasoning, LLMs have progressed
from solving elementary word problems to demonstrating automated theorem-proving capabilities
and achieving state-of-the-art performance on competitive, Olympiad-level benchmarks
\citep{castelvecchi2025ai,wei2022chain,trinh2024solving,shao2024deepseekmath}. This capability
extends to time-series analysis, where foundation models serve as generalist forecasters for
complex temporal dynamics in weather, traffic, and energy consumption
\citep{jin2023time,zhou2023one,chang2025llm4ts,rasul2023lag}. The financial sector has
likewise witnessed deployment of specialized LLMs for quantitative trading, risk assessment,
and earnings report analysis \citep{wu2023bloomberggpt,liu2023fingpt}. In industrial and
scientific applications, agents are increasingly utilized for material discovery, chemical
synthesis planning, and manufacturing process optimization
\citep{m2024augmenting,taylor2022galactica,boiko2023emergent}.

Despite the semantic diversity of these fields, they share a critical unifying foundation:
reliance on numerical data processing. Whether interpreting stock tickers, sensor readings,
timestamp intervals, or algebraic coefficients, the ability to accurately parse, manipulate,
and generate numerical tokens is a prerequisite for reliable performance.

Yet a jarring contradiction persists: while LLMs demonstrate sophisticated conceptual
understanding in high-level domains, their intrinsic ability to process the underlying
numerical data remains fundamentally unreliable and prone to hallucination
\citep{dziri2023faith}. This phenomenon---often termed a lack of \emph{numeracy} or
\emph{number sense}---represents a fundamental disconnect between linguistic fluency and
computational competence. State-of-the-art models frequently hallucinate on elementary
magnitude comparisons (e.g., asserting $9.11 > 9.9$), fail to maintain precision when adding
large integers, and exhibit extreme sensitivity to surface-level changes in problem phrasing
\citep{mirzadeh2024gsm}. These are not isolated edge cases; they are systemic failures that
suggest current LLMs process numbers as \emph{semantic tokens} rather than as \emph{mathematical
values}.

This discrepancy is not merely an academic curiosity; it constitutes a critical reliability
bottleneck for any real-world application. Complex mathematical reasoning necessarily rests on
the bedrock of reliable elementary arithmetic. When models hallucinate intermediate numerical
results, even a flawless logical deduction chain leads to an incorrect final answer---a
``weakest link'' problem that renders the model untrustworthy for autonomous tasks.

\textbf{Thesis: numeracy failures as grounding failures.}
We argue that these failures arise from \emph{insufficient numerical grounding}---the inability
to reliably map numeral tokens to their mathematical values and operate on those values with
algorithmic fidelity. Building on Harnad's (1990) symbol grounding problem
\citep{harnad1990symbol} and Dehaene's (2011) cognitive theory of number sense
\citep{dehaene2011number}, we formalize this argument in the \emph{Numerical Grounding
Framework} (NGF). NGF decomposes numerical competence into two complementary dimensions:
\emph{Representational Grounding} (RG), the faithful mapping between a numeral's surface form
and an internal representation that preserves ordering, magnitude, and format equivalence; and
\emph{Procedural Grounding} (PG), the faithful execution of arithmetic procedures whose outputs
are consistent with the mathematical definition of those operations. We use NGF as the
organizing lens for every claim in the paper: each failure mode, each architectural root cause,
and each mitigation strategy is mapped to RG, PG, or both.

Despite the critical role of numeracy, no existing survey has placed a primary, dedicated
focus on the intrinsic numerical ability of LLMs as distinct from high-level mathematical
reasoning \citep{ahn2024large}. Foundational issues of numerical primitives---tokenization
failures, place-value recognition, and algorithmic generalization over length---have been
consistently treated as secondary consequences of poor reasoning rather than primary,
independent weaknesses requiring dedicated diagnosis and architectural solutions. This
oversight motivates the present survey.

The main contributions of this survey are as follows.

\textbf{1. Theoretical Framework (Section~\ref{sec3}).}
We introduce NGF, an original two-part construct grounded in cognitive science
\citep{dehaene2011number,spelke2007core} and in the symbol grounding tradition
\citep{harnad1990symbol}, and applied consistently throughout the paper to predict which
failure modes respond to which interventions.

\textbf{2. Comprehensive Related Work (Section~\ref{sec2}).}
We survey eight strands of literature that intersect with LLM numeracy: number probing and
internal representation studies, emergent abilities and scaling laws, alternative tokenization
strategies, compositional arithmetic, multilingual numeracy, process-reward models,
application-domain failure case studies, and cognitive-science baselines.

\textbf{3. Evaluation Landscape (Section~\ref{sec3}).}
We critically assess the shift from saturated legacy benchmarks (GSM8K, MATH) to the new
generation of specialized diagnostic benchmarks---NumericBench, Number Cookbook, GSM-Symbolic,
GSM-Ranges, MGSM, and BIG-Bench Hard.

\textbf{4. Systematic Taxonomy and Failure Mode Analysis (Section~\ref{sec3}).}
We present a taxonomy of numerical tasks across four domains, annotated with grounding type
(RG/PG) and characteristic failure modes. We identify four primary failure classes: Fragility,
Tokenization Artifacts, Length Generalization Failure, and Algorithmic Asymmetry.

\textbf{5. Root Cause Analysis (Section~\ref{sec4}).}
We provide a deep analysis of four structural root causes mapped onto NGF: BPE tokenization
(damages RG), positional encodings (damages PG), embedding discontinuity (damages RG), and
pretraining data distribution (damages both).

\textbf{6. Coordinated Empirical Evaluation (Section~\ref{sec5}).}
We operationalize NGF across Number Cookbook, NumericBench, and GSM-Symbolic, evaluating
frontier model families under both direct-answer and extended-reasoning regimes. The results
test four predictions: RG--PG dissociation, reasoning compensation, tokenizer-specific RG
profiles, and primitive--contextual transfer.

\textbf{7. Mitigation Strategies (Section~\ref{sec6}).}
We systematically evaluate proposed mitigation strategies organized by the grounding dimension
they address (RG, PG, or both), and characterize the \emph{Pretrained-Model Constraint}:
architectural and tokenization-level interventions that dramatically improve scratch-trained
models are frequently inapplicable on pretrained models.

\textbf{8. Future Directions and Deployment Recommendations (Section~\ref{sec7}).}
We synthesize current limitations into a research agenda and provide concrete deployment
recommendations for practitioners.

\section{Related Work}
\label{sec2}

The literature relevant to LLM numeracy spans eight intersecting strands, each contributing
partial insight that the NGF lens unifies. We survey them in turn before positioning the
present survey relative to prior work.

\subsection{Surveys on LLM Mathematical Reasoning}

\citet{ahn2024large} provide a comprehensive review of progress and challenges in LLM
mathematical reasoning, covering Chain-of-Thought prompting, tool use, and benchmark
evaluation. Their focus, however, remains on high-level problem-solving; the structural causes
of elementary numerical failure receive no dedicated treatment.
\citet{frieder2023mathematical} empirically evaluate the mathematical capabilities of GPT-3.5
and GPT-4 across a wide range of topics, identifying consistent failure patterns. While their
evaluation surfaces numerical errors, these are treated as symptoms of broader reasoning
limitations rather than as structural phenomena. Neither survey adopts an RG/PG decomposition
of numerical competence.

\subsection{Number Probing and Internal Representation}

A growing literature applies mechanistic interpretability tools to the question of whether LLMs
internally represent numerical magnitude as a continuous quantity.
\citet{stolfo2023mechanistic} provide a mechanistic decomposition of arithmetic circuits in
transformer LLMs, identifying specific attention heads and MLP layers that participate in
addition tasks. \citet{gurnee2023language} demonstrate that language models maintain internal
linear representations of spatial and temporal magnitudes that can be decoded from the residual
stream. These results suggest that some degree of representational grounding emerges
spontaneously at scale, but the geometry remains noisy, format-sensitive, and entangled with
surface tokenization---motivating the structural concerns explored in Section~\ref{sec4}.

\subsection{Emergent Abilities and Scaling}

\citet{wei2022emergent} document discontinuous jumps in benchmark accuracy at particular model
scales, including arithmetic capabilities. \citet{srivastava2022beyond} introduce BIG-Bench, a
large collaborative benchmark containing numerous numerical primitive tasks (digit manipulation,
modular arithmetic, magnitude comparison), and BIG-Bench Hard isolates the subset on which scale
alone fails to deliver reliable performance. These findings reveal that scaling improves
\emph{some} numerical capabilities while leaving others structurally fragile---a pattern that
NGF clarifies by distinguishing RG capabilities (which appear to benefit modestly from scale)
from PG capabilities (which often plateau without architectural or data-distribution intervention).

\subsection{Arithmetic and Alternative Tokenization}

A targeted literature examines arithmetic directly in neural architectures. Working before the
era of large pretrained models, \citet{nogueira2021investigating} systematically demonstrate
length generalization failures and tokenization sensitivity in transformer models on simple
arithmetic tasks. \citet{charton2022transformers} studies conditions under which transformers
can learn mathematical functions such as greatest common divisors. More recently, Llama~3
\citep{dubey2024llama} adopts a fixed three-digit tokenization scheme that abandons BPE for
numerals. The design is consistent with the broader evidence reviewed in this paper:
tokenizer choices measurably shape RG performance on magnitude comparison, format conversion,
and digit-aware tasks, though such benefits require vocabulary and pretraining decisions made
before a model is trained.

\subsection{Compositional Arithmetic and Length Generalization}

\citet{dziri2023faith} study the limits of transformers on compositional tasks, demonstrating
that performance degrades systematically as reasoning chains lengthen. \citet{anil2022exploring}
characterize the length generalization failure across multiple tasks and architectures, showing
it to be a robust pattern rather than an artifact of specific training data. Numerical
computation is a special case of compositionality---carrying digits, aligning decimal
points---where the same structural failures manifest as PG breakdowns under increasing operand
length.

\subsection{Multilingual Numeracy}

Numerical reasoning research has historically focused on English-only inputs.
\citet{shi2022language} introduce MGSM, a multilingual translation of GSM8K covering ten
languages with varied numeral conventions (digit grouping characters, decimal separators,
native-script digits). Performance gaps between English and other languages exceed those
observed for non-numerical tasks, indicating that surface-form variation in numerical notation
interacts with tokenization to compound RG failure. Cross-lingual numeracy is a critical but
understudied gap addressed in our future-directions agenda.

\subsection{Process Reward Models}

Standard reinforcement learning approaches to mathematical reasoning reward only the final
answer (outcome supervision). \citet{lightman2023lets} and \citet{uesato2022solving}
demonstrate that \emph{process reward models}, which provide a reward signal at each
intermediate reasoning step, yield substantially more reliable arithmetic execution than
outcome-only training. These results suggest that PG can be improved without architectural
changes by changing the training signal---an approach orthogonal to RG-focused tokenizer
interventions.

\subsection{Application-Domain Failure Cases}

Beyond benchmark numbers, real-world failure cases sharpen the practical stakes of LLM
numeracy. Clinical decision support has been documented to fail on dose-comparison tasks
where a misread decimal could be fatal. Financial summarization systems have produced
material misstatements by miscopying or mis-aggregating numerical entries from earnings
reports. These domain-specific failure cases motivate the deployment recommendations of
Section~\ref{sec6_practical}, particularly the use of Tool-Use scaffolding for
high-reliability contexts.

\subsection{Cognitive Science Baselines}

The Numerical Grounding Framework is informed by, and intended to be commensurable with,
classical results in cognitive science. \citet{harnad1990symbol}'s \emph{symbol grounding
problem} observes that purely symbol-symbol systems cannot acquire semantics without some link
to extrasymbolic referents---a concern that translates directly to the question of how LLMs
link numeral tokens to magnitude. \citet{dehaene2011number} reviews behavioral and neural
evidence that humans possess an \emph{approximate number system} organized along a logarithmic
mental number line, supplemented by an exact small-number system. \citet{spelke2007core}
characterize core knowledge of number as one of a small number of evolutionarily conserved
cognitive systems present from infancy. NGF treats RG as the LLM analogue of approximate
number representation and PG as the analogue of exact symbolic computation.

\subsection{Positioning of This Survey}

The present survey is distinguished from prior work in four respects. First, we introduce NGF
as an original analytical framework that explicitly separates representational from procedural
numerical competence, providing a unified vocabulary for organizing failure modes, benchmarks,
and mitigations. Second, we provide systematic coverage of the 2024--2025 diagnostic benchmark
generation. Third, we evaluate mitigation strategies from the perspective of \emph{pretrained-model
compatibility}---a practical constraint that prior work has not systematically addressed.
Fourth, we connect the LLM numeracy literature to classical cognitive science via Harnad's
symbol grounding tradition and Dehaene's number-sense framework.

\section{The Numerical Grounding Framework}
\label{sec3}

To address the numeracy gap, we must first decouple high-level mathematical reasoning from
fundamental number sense. This section formalizes the Numerical Grounding Framework, documents
the survey methodology, surveys the new generation of diagnostic benchmarks, and categorizes
the systematic failure modes exhibited by current state-of-the-art models.

\subsection{Defining Numerical Grounding (NGF)}
\label{sec3_ngf}

In human cognition, \emph{numeracy} is distinct from \emph{mathematics}. While mathematics
involves abstract reasoning, logic, and proof, numeracy is the literacy of numbers---the
ability to read, represent, compare, and manipulate quantities. \citet{harnad1990symbol}'s
symbol grounding problem identifies the central difficulty: a system that manipulates only
symbols without linking them to extrasymbolic referents cannot acquire the meanings of those
symbols. \citet{dehaene2011number} characterizes human number competence as the joint operation
of two systems: an approximate number system that maps numerals onto a continuous magnitude
representation, and an exact symbolic system that executes procedures over digits. NGF adapts
this two-system perspective to language models.

\textbf{Definition.} A language model exhibits \emph{numerical grounding} when it satisfies
the conjunction of two properties:

\begin{itemize}
\item \textbf{Representational Grounding (RG).} The model maps the surface form of a numeral
to an internal representation that preserves (i)~ordering with respect to magnitude,
(ii)~magnitude itself, and (iii)~format equivalence across alternative surface forms of the
same value (e.g., $0.5 \equiv \tfrac{1}{2} \equiv 5 \times 10^{-1} \equiv 50\%$).

\item \textbf{Procedural Grounding (PG).} The model executes arithmetic procedures whose
outputs are consistent with the mathematical definition of those operations, including
correctness on inputs whose length or magnitude exceeds those observed during training.
\end{itemize}

A model with RG can recognize that two surface forms denote the same quantity and that one
quantity is greater than another, but does not necessarily compute correctly with them. A
model with PG can mechanically execute an algorithm whose inputs and outputs it nonetheless
treats as opaque symbols. Robust numerical competence requires both.

\textbf{Mapping of failure modes to NGF.}
The four failure modes documented in Section~\ref{sec3_failure} map onto NGF as follows. The
Fragility failure (sensitivity to surface number choice and to irrelevant distractors) is
primarily an RG failure: the model lacks a semantic schema for which numbers are causally
relevant. The Tokenization Artifact failure (e.g., $9.11 > 9.9$) is an RG failure: the
surface-to-representation mapping is non-monotonic. The Length Generalization failure is a
PG failure: the procedure does not extend beyond trained operand lengths. The Algorithmic
Asymmetry failure (subtraction sign blindness, division weakness) is a PG failure: the
procedure is incompletely learned.

\begin{figure}[t]
\centering
\includegraphics[width=0.96\linewidth]{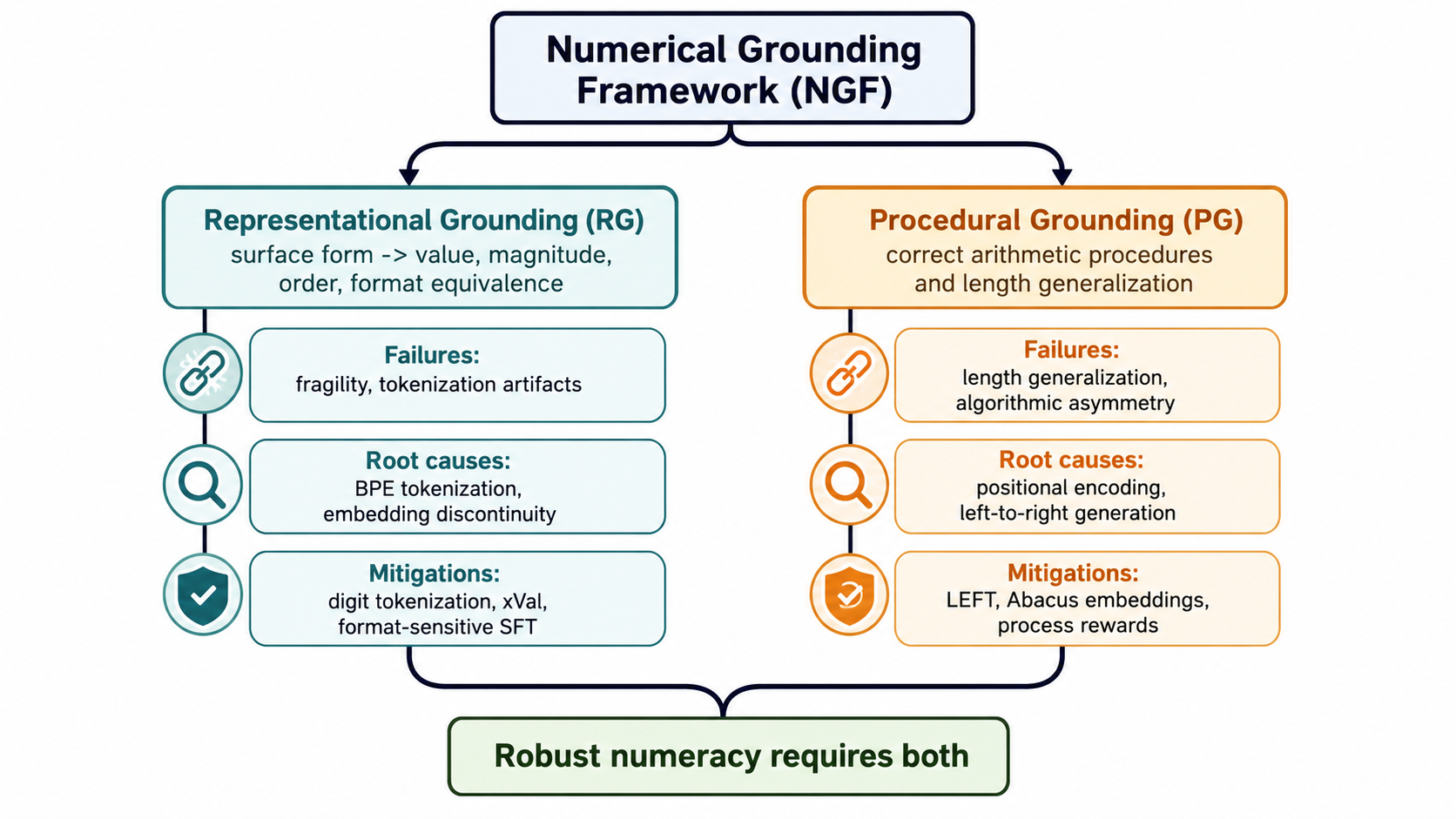}
\caption{\textbf{Numerical Grounding Framework (NGF).} The framework decomposes LLM numeracy
into Representational Grounding (RG), which concerns surface-form-to-value mapping, and
Procedural Grounding (PG), which concerns faithful execution of arithmetic procedures.
This separation provides the organizing principle for the benchmark taxonomy, root-cause
analysis, empirical results, and mitigation strategies in the remainder of the paper.}
\label{fig:ngf-framework}
\end{figure}

\subsection{Survey Methodology}
\label{sec3_methodology}

This survey follows a structured narrative review methodology appropriate for a rapidly
evolving area where systematic meta-analytic protocols have not yet been standardized.
Literature was identified through three complementary channels: (1)~forward and backward
citation tracing from the four primary 2024--2025 diagnostic benchmark papers (NumericBench,
Number Cookbook, GSM-Symbolic, GSM-Ranges); (2)~keyword search on arXiv (cs.CL, cs.LG); and
(3)~targeted review of NeurIPS, ICLR, ICML, ACL, and EMNLP proceedings from 2021--2025 for
papers addressing arithmetic or numerical evaluation in autoregressive language models.
Papers addressing purely symbolic mathematics, vision-language numeracy without a
language-model component, or downstream reasoning tasks without numerical primitive analysis
were excluded.

\subsection{The New Evaluation Landscape}
\label{sec3_benchmarks}

Traditional benchmarks such as GSM8K \citep{cobbe2021gsm8k} and MATH
\citep{hendrycks2021math} have become saturated, with frontier models achieving $>95\%$
accuracy on GSM8K. However, this high performance often masks underlying fragility: models
may solve standard problems through memorized solution templates while failing on modest
reformulations. To diagnose grounding failures specifically, a generation of targeted
diagnostic benchmarks emerged in 2024--2025.

\begin{figure}[t]
\centering
\includegraphics[width=0.88\linewidth]{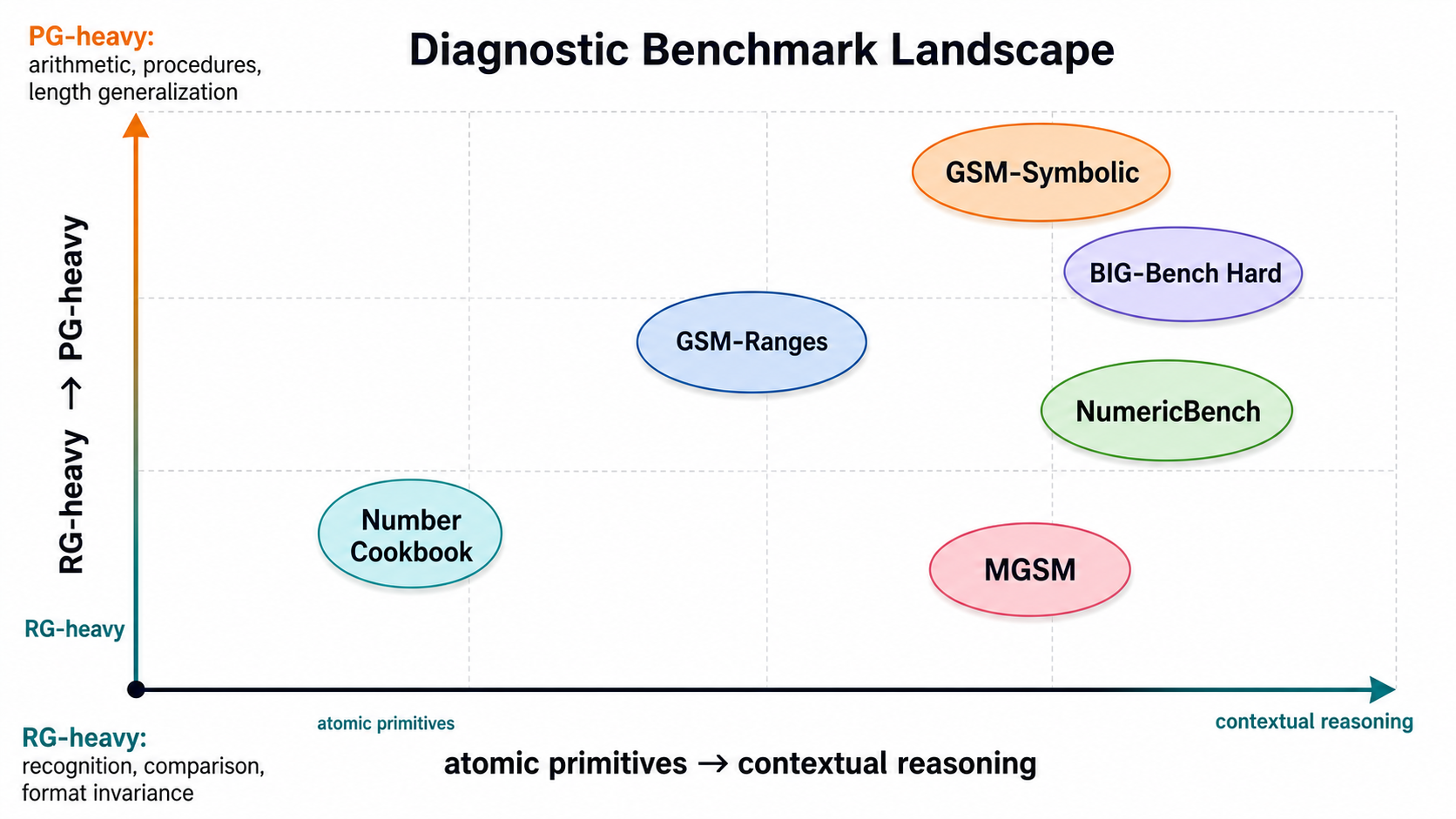}
\caption{\textbf{Diagnostic benchmark landscape.} Current numeracy benchmarks occupy different
regions of the space from atomic numerical primitives to contextual word-problem reasoning,
and from RG-heavy recognition/comparison tasks to PG-heavy procedural arithmetic. The spread
is essential: no single benchmark is sufficient to characterize numerical grounding.}
\label{fig:benchmark-landscape}
\end{figure}

\textbf{NumericBench}~\citep{li2025numericbench} is a comprehensive suite evaluating six
fundamental numerical abilities across real-world data and synthetic lists. Its distinctive
contribution is testing \emph{in-context} numeracy: finding and operating on numbers embedded
within natural language passages rather than in isolated equations. This makes NumericBench
uniquely sensitive to RG.

\textbf{Number Cookbook}~\citep{wang2024number} is a diagnostic testbed isolating 17 atomic
numerical tasks across four distinct numerical representations---integer, float, fraction, and
scientific notation. Its fine-grained decomposition makes it especially useful for identifying
which specific tokenizer-induced structural deficits produce which failure types.

\textbf{GSM-Symbolic and GSM-NoOp}~\citep{mirzadeh2024gsm} generate thousands of symbolic
variations of GSM8K problems by substituting different numerical values (measuring
\emph{fragility}) and by inserting irrelevant numerical distractors (measuring
\emph{distraction resistance}).

\textbf{GSM-Ranges} extends the GSM8K framework by systematically varying the magnitude of
numbers across several orders of magnitude, directly measuring length generalization---a PG
failure documented in Section~\ref{sec3_failure}.

\textbf{MGSM}~\citep{shi2022language} extends GSM8K to ten languages, providing the only
widely used multilingual probe of arithmetic word-problem reasoning. The English-to-other-language
performance gap on MGSM exceeds that on non-numerical multilingual benchmarks.

\textbf{BIG-Bench and BIG-Bench Hard}~\citep{srivastava2022beyond} include numerous numerical
primitive tasks that have proven resistant to scale and prompting; BIG-Bench Hard isolates the
subset on which performance remains low even at frontier scale.

\textbf{Limitations of the New Benchmarks.}
These benchmarks collectively advance grounding measurement but share several important gaps.
They are primarily English-language; MGSM is the principal exception but covers only word
problems. Performance on Number Cookbook atomic tasks does not reliably predict performance on
NumericBench in-context tasks, suggesting that RG-primitive and RG-contextual capabilities are
partially dissociated. Finally, no benchmark yet covers the full distributional range of
numerical formats encountered in real deployments.

\subsection{Taxonomy of Numerical Tasks}
\label{sec3_taxonomy}

Synthesizing these benchmarks, we categorize the numeracy landscape into four primary task
domains. Table~\ref{tab:numerical-tasks} annotates each category with its primary grounding
type (RG~=~Representational, PG~=~Procedural), which predicts which root causes
(Section~\ref{sec4}) and mitigation strategies (Section~\ref{sec6}) are most relevant.

\begin{table}[t]
\centering
\caption{Taxonomy of numerical reasoning task categories with grounding type annotation.
RG~=~Representational Grounding, PG~=~Procedural Grounding.}
\label{tab:numerical-tasks}
\setlength{\tabcolsep}{4pt}
\begin{tabular}{llcp{5.8cm}}
\toprule
\textbf{Domain} & \textbf{Task Category} & \textbf{NGF} & \textbf{Definition and Example} \\
\midrule
\multirow{4}{*}{Representation}
  & Recognition \& Extraction & RG
    & Identifying numerical entities in unstructured text. \\
  & & & \textit{``Extract the Q3 revenue from the report.''} \\[4pt]
  & Format Conversion & RG
    & Converting between formats. \textit{``Convert 3/4 to a decimal.''} \\
\midrule
\multirow{4}{*}{Arithmetic}
  & Elementary Operations & PG
    & The four basic operations ($+$, $-$, $\times$, $\div$) and modulus. \\
  & & & \textit{``Calculate $1234 \times 5678$.''} \\[4pt]
  & Length Generalization & PG
    & Performing operations on inputs longer than training examples. \\
  & & & \textit{``Add these two 50-digit integers.''} \\
\midrule
\multirow{4}{*}{Comparison}
  & Magnitude Comparison & RG
    & Determining ordering relationships by value.
      \textit{``Is 9.11 greater than 9.9?''} \\[4pt]
  & Sorting & RG
    & Ordering a list of values correctly. \\
\midrule
\multirow{4}{*}{Structural}
  & Digit Manipulation & PG
    & Accessing or modifying specific digits. \\
  & & & \textit{``What is the 3rd digit of 3.1415?''} \\[4pt]
  & Contextual Retrieval & RG
    & Locating specific values in long-context tables. \\
\bottomrule
\end{tabular}
\end{table}

\subsection{Analysis of Failure Modes}
\label{sec3_failure}

Evaluation on these benchmarks reveals that LLM numeracy failures are not random; they are
systematic and structural. We identify four primary failure modes, each linked to its
grounding type and root cause.

\subsubsection{The Fragility Failure (Pattern Matching vs.\ Reasoning) --- RG}

The GSM-Symbolic study \citep{mirzadeh2024gsm} revealed that model performance is highly
sensitive to the specific numbers used in a problem. Simply changing the values in a question
can cause significant accuracy drops, suggesting that models rely on memorized solution
templates. Even more severe is the NoOp effect: adding irrelevant numerical information causes
performance to drop by up to 65\% in some models. Models struggle to filter numerical
``noise'' from ``signal.'' This is a \textbf{Representational Grounding (RG)} failure: the
model lacks a semantic schema for which numbers are causally relevant.

\subsubsection{The Tokenization Artifact Failure --- RG}

One of the most pervasive grounding failures is the inability to correctly compare decimal
magnitudes, exemplified by models asserting $9.11 > 9.9$.

Standard BPE tokenizers do not split numbers according to mathematical logic; they split
according to co-occurrence frequency in the training corpus. The decimal $9.11$ is often
tokenized as $[\texttt{9}, \texttt{.}, \texttt{11}]$, while $9.9$ tokenizes as
$[\texttt{9}, \texttt{.}, \texttt{9}]$. In token-id space, where no continuous magnitude is
encoded, the model compares the final tokens and incorrectly concludes $9.11 > 9.9$. The
tokenization of any given number is also context-dependent. The result is a non-monotonic
surface-to-value mapping---the canonical failure of \textbf{RG}.

\subsubsection{The Length Generalization Failure --- PG}

LLMs exhibit a sharp performance cliff as the length of number operands increases beyond the
range seen during training. Both Number Cookbook \citep{wang2024number} and studies of Abacus
Embeddings \citep{mcleish2024transformers} show that models trained on addition of up to $N$
digits fail catastrophically when presented with $N{+}1$ or more digits.
\citet{anil2022exploring} document the same pattern across diverse compositional tasks. The
mechanism is rooted in positional encoding: standard encodings such as RoPE
\citep{su2022roformer} struggle to extrapolate to token distances unseen during training. This
is a pure \textbf{Procedural Grounding (PG)} failure.

\subsubsection{The Algorithmic Asymmetry (Subtraction and Division Gap) --- PG}

For subtraction when $A < B$, models frequently produce the correct magnitude but omit the
leading negative sign---a ``sign blindness.'' Division remains the hardest elementary operation,
as it requires iterative estimation and multiplication that is difficult to simulate in a single
forward pass without extensive Chain-of-Thought scaffolding. Both are \textbf{PG} failures: the
algorithm is not faithfully internalized.

\subsection{From Taxonomy to Evidence}
\label{sec3_performance}

The taxonomy above is not merely descriptive. It makes empirical predictions that can be
checked across model families and benchmark ecologies. If RG and PG are genuinely distinct,
then models should display different strength profiles across comparison, format-conversion,
digit-access, arithmetic, and length-generalization tasks. If tokenization is a structural RG
bottleneck, different tokenizer implementations should create different ``blind spots'' even
when the underlying mathematical task is identical. If reasoning primarily supplies procedural
scaffolding, then extended reasoning should help PG more than RG, especially out of domain.

Section~\ref{sec5} operationalizes these predictions through a coordinated evaluation on
Number Cookbook, NumericBench, and GSM-Symbolic. The resulting figures are placed in the
empirical section rather than here so that the conceptual story (framework, benchmarks,
failure modes, root causes) remains cleanly separated from the evidence that tests it.

\FloatBarrier

\section{Root Causes of Innumeracy}
\label{sec4}

The failures described above are not inexplicable; they are the predictable result of specific
architectural and data-distribution choices in current LLMs. The literature, viewed through
the NGF lens, identifies four primary structural sources of innumeracy: BPE tokenization
(damaging RG), positional encoding (damaging PG), embedding discontinuity (damaging RG), and
pretraining data distribution (damaging both).

\begin{figure}[t]
\centering
\includegraphics[width=0.95\linewidth]{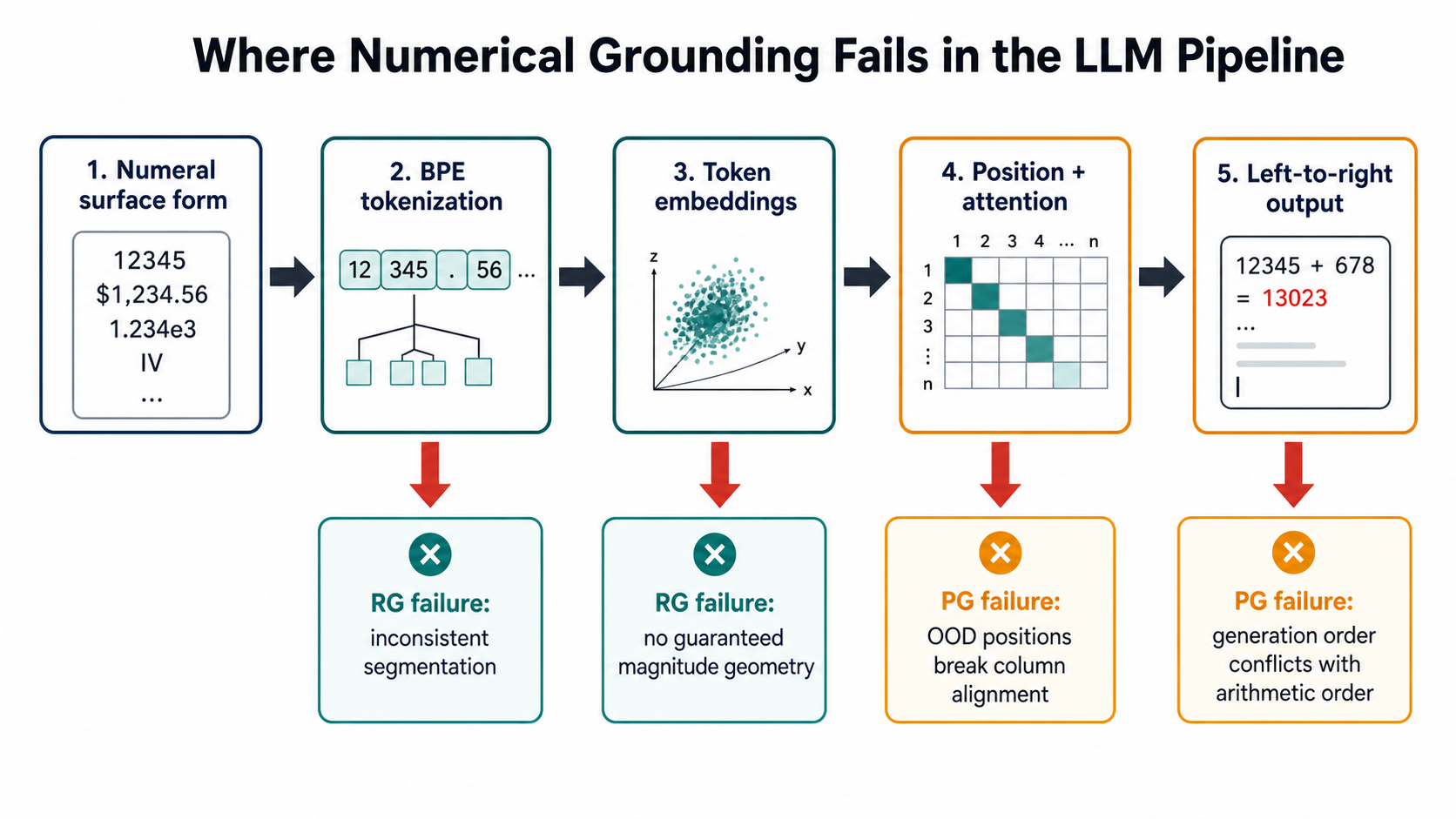}
\caption{\textbf{Where numerical grounding fails in the LLM pipeline.} Numeral strings pass
through tokenizer segmentation, embedding lookup, positional attention, and left-to-right
decoding. Each stage introduces a different failure channel: segmentation and embedding
geometry primarily damage RG, while positional extrapolation and autoregressive generation
order primarily damage PG.}
\label{fig:pipeline-failures}
\end{figure}

\subsection{The Tokenization Bottleneck: Byte-Pair Encoding (BPE) --- Damages RG}
\label{sec4_bpe}

The most significant barrier to RG is the Byte-Pair Encoding (BPE) algorithm used to tokenize
text before processing. BPE compresses text by iteratively merging the most frequent adjacent
byte pairs into single tokens. While highly efficient for natural language, this
frequency-driven merging is \emph{mathematically destructive} for numbers
\citep{nogueira2021investigating}.

\subsubsection{Mechanism of Failure}

\textbf{Inconsistent segmentation.} A number such as $12345$ may tokenize as a single token
$[\texttt{12345}]$ if it appears frequently enough in training data, while $12346$ tokenizes
as $[\texttt{123}, \texttt{46}]$ because that boundary happens to be the highest-frequency
merge. The model sees these as completely different sequences of ``words,'' making it
impossible to learn a consistent place-value rule. The same failure underlies the
$9.11 > 9.9$ error.

\textbf{Right-to-left dependency with left-to-right generation.} Standard arithmetic on
integers proceeds from right to left; LLMs generate tokens left to right. BPE tokenization
further obscures individual digit positions, contributing to PG breakdown on multi-digit
addition and multiplication.

\textbf{Context-dependent tokenization.} The same numeral may tokenize differently depending
on surrounding characters, compounding the inconsistency that prevents learning a stable
numerical schema.

\begin{figure}[t]
\centering
\includegraphics[width=0.88\linewidth]{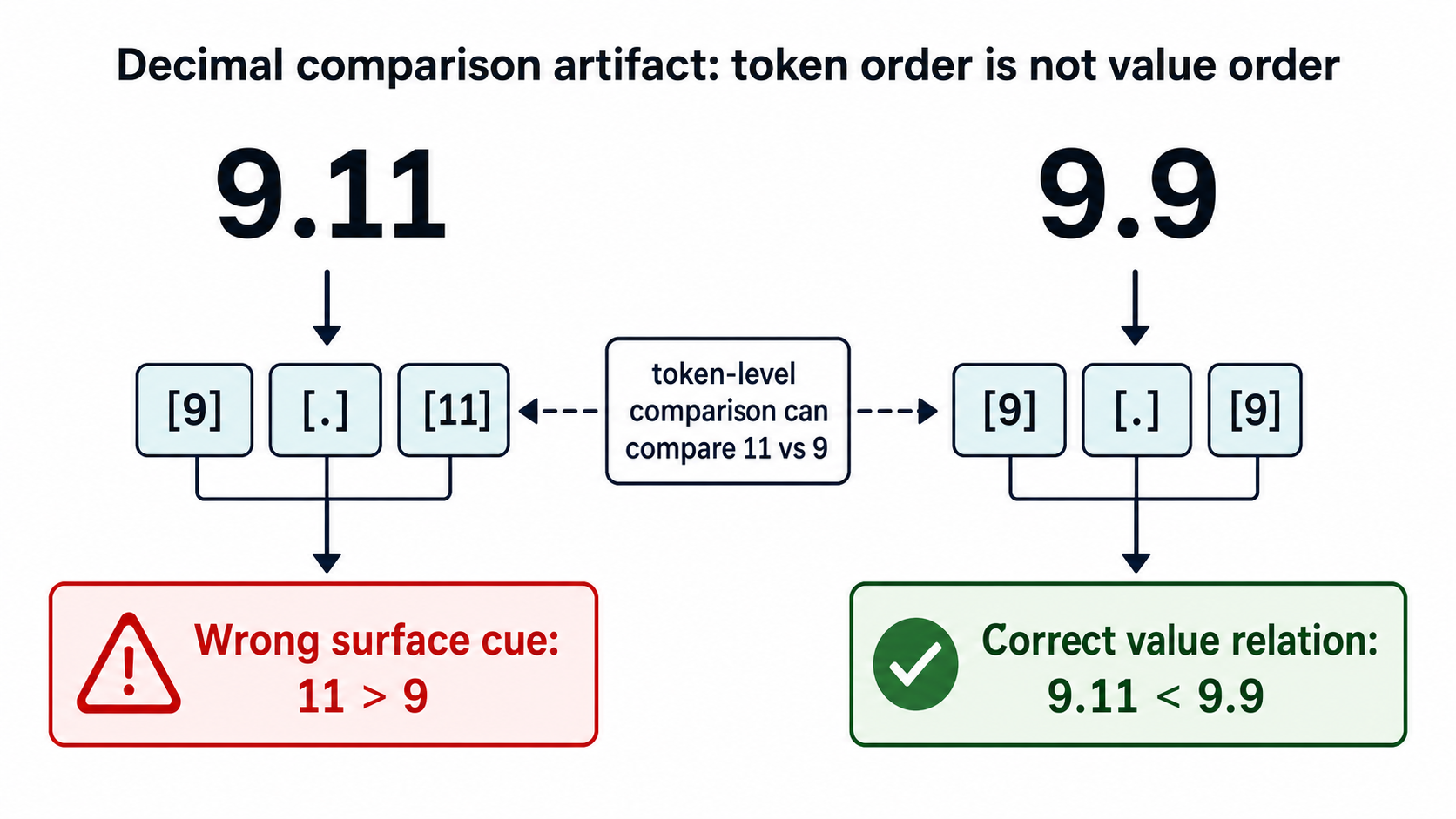}
\caption{\textbf{Decimal comparison as a tokenization artifact.} A mathematically simple
comparison such as $9.11 < 9.9$ can be corrupted when tokenization exposes the suffix
\texttt{11} as a token comparable to \texttt{9}. The model is nudged toward an invalid
surface-level cue rather than a value-preserving representation.}
\label{fig:decimal-token-artifact}
\end{figure}

\subsection{Positional Encoding and Length Generalization --- Damages PG}
\label{sec4_pe}

Transformers process all tokens in parallel and rely on Positional Encodings to understand
token order. The choice of positional encoding largely determines a model's ability to
generalize to longer number sequences than it was trained on, making positional encoding the
principal driver of \textbf{PG} failure under length extension.

\subsubsection{RoPE and Its Limitations}

Rotary Positional Embeddings (RoPE), used in Llama and most modern LLMs \citep{su2022roformer},
encode position by rotating the query and key vectors in attention. RoPE works well within the
training distribution but struggles on out-of-distribution lengths: rotation frequencies for
larger positional distances were never encountered during training. Position Interpolation and
YaRN \citep{peng2023yarn} address context-window extension but their benefit for arithmetic
length generalization is less well-studied.

\subsubsection{ALiBi and Recency Bias}

ALiBi \citep{press2022alibi} adds a distance-proportional penalty to attention logits, biasing
attention toward nearby tokens. While ALiBi generalizes better than sinusoidal embeddings for
language, this recency bias is harmful for arithmetic: carry propagation requires long-range
dependency between the least significant digit and every more significant digit. ALiBi's
distance penalty suppresses exactly this necessary dependency.

\subsection{Embedding Space Discontinuity --- Damages RG}
\label{sec4_embedding}

Humans understand numbers as lying on a continuous ordered line; an LLM's token embedding
layer treats each number token as a discrete categorical entry. The token \texttt{10} is
represented with no guaranteed geometric relationship to \texttt{11} or \texttt{9}. There is
no inherent inductive bias in the embedding layer that encodes $10 < 11$.

The xVal framework \citep{golkar2023xval} provides empirical evidence for this discontinuity:
when standard LLM number token embeddings are projected via PCA, the projections of successive
integers do not form a monotonically ordered sequence. Probing studies
\citep{stolfo2023mechanistic,gurnee2023language} refine this picture, showing that magnitude is
partially decodable from deeper hidden states but is not cleanly localized at the embedding
layer.

\subsection{Pretraining Data Distribution --- Damages Both RG and PG}
\label{sec4_data}

A fourth structural cause, largely overlooked in earlier analyses, is the distribution of
numerical content in pretraining corpora. Numerical sequences are sparse in natural text
compared to natural-language tokens, and the distribution that does exist is skewed: round
numbers, common dates, and small integers dominate, while long multi-digit operands,
irrational decimals, and scientific-notation forms are vastly underrepresented.
\citet{srivastava2022beyond} provide empirical evidence that arithmetic accuracy correlates
with operand frequency in pretraining data far more strongly than with model scale.

This distributional sparsity damages both grounding dimensions. \textbf{RG} suffers because
rare numerical surface forms are under-trained: their tokenizations are inconsistent, their
embeddings are noisy, and the model has fewer examples from which to learn a stable
surface-to-value mapping. \textbf{PG} suffers because step-by-step demonstrations of
arithmetic procedures are scarce in unstructured web text; synthetic data generation or
process-reward training is needed to close the gap.


\section{Empirical Evaluation}
\label{sec5}

This section operationalizes NGF through a coordinated evaluation across three complementary
benchmarks: Number Cookbook (atomic numerical primitives), NumericBench (contextual numerical
reasoning), and GSM-Symbolic (parametric fragility). The goal is not simply to report another
leaderboard, but to test whether the RG/PG decomposition predicts observed model behavior.

\subsection{Motivation and Research Questions}
\label{sec5_rqs}

The evaluation is structured around four questions derived from the framework:

\begin{itemize}
\item \textbf{RQ1 (Dissociation).} Are RG and PG empirically dissociable across models, and
is RG consistently easier than PG?
\item \textbf{RQ2 (Reasoning compensation).} Does extended reasoning close PG gaps more than
RG gaps, especially on out-of-domain inputs?
\item \textbf{RQ3 (Tokenization contrast).} Do models with different tokenizers exhibit
systematically different RG profiles on matched tasks?
\item \textbf{RQ4 (Primitive--contextual transfer).} Does atomic-task performance predict
contextual numerical reasoning, or are primitive and contextual numeracy partially dissociated?
\end{itemize}

\subsection{Experimental Setup}
\label{sec5_setup}

\textbf{Model set.} We evaluate three frontier model families: GPT-5.4, Claude Opus~4.6, and
Gemini~3. Gemini is evaluated under two \texttt{thinkingLevel} settings---MINIMAL and HIGH---to
isolate the effect of extended internal reasoning. All models are accessed through a unified API
proxy with temperature~0.

\textbf{Benchmark 1: Number Cookbook.} All 44 task--representation combinations are evaluated
and grouped into an RG battery (magnitude comparison, format conversion, significant-figure
rounding, digit counting, and number length) and a PG battery (arithmetic operations,
digit-wise operations, and digit extraction). Inputs are split into in-domain and out-of-domain
operand lengths.

\textbf{Benchmark 2: NumericBench.} We sample contextual numerical tasks spanning arithmetic,
context arithmetic, different-digit arithmetic, numerical list comprehension, and mixed
number-string extraction. This benchmark tests whether isolated primitive competence transfers
to naturalistic text and table contexts.

\textbf{Benchmark 3: GSM-Symbolic.} We evaluate the main split and added-clause variants,
measuring both accuracy and per-template fragility under numerical substitution.

\subsection{Aggregate Results and RG--PG Dissociation}
\label{sec5_dissociation}

\begin{table}[t]
\centering
\caption{Aggregate Number Cookbook accuracy. Exact-match accuracy collapses out of domain,
while digit-match accuracy reveals partial procedural structure even when final answers are
wrong.}
\label{tab:nc-aggregate}
\setlength{\tabcolsep}{5pt}
\begin{tabular}{lcccc}
\toprule
\textbf{Model} & \textbf{Exact (in)} & \textbf{Exact (out)} & \textbf{Digit (in)} & \textbf{Digit (out)} \\
\midrule
GPT-5.4          & 76.6 & 40.2 & 89.5 & 65.2 \\
Claude Opus 4.6  & 82.2 & 50.0 & 88.2 & 62.0 \\
Gemini 3 MINIMAL & 92.5 & 63.4 & --   & --   \\
Gemini 3 HIGH    & 86.4 & 85.0 & --   & --   \\
\bottomrule
\end{tabular}
\end{table}

\begin{figure}[t]
\centering
\includegraphics[width=0.95\linewidth]{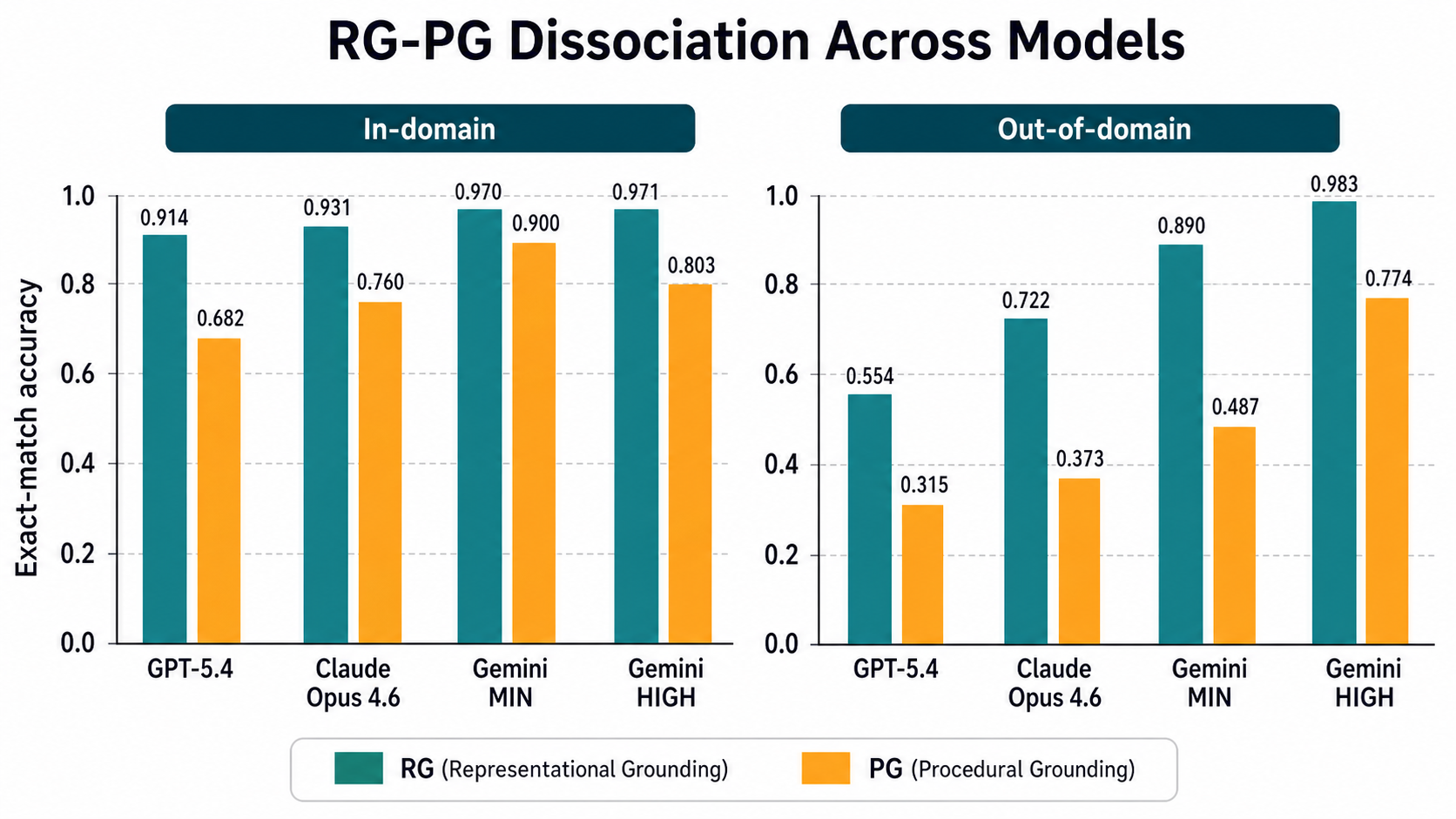}
\caption{\textbf{RG--PG dissociation.} Across all model configurations, RG accuracy exceeds PG
accuracy in both in-domain and out-of-domain regimes. The gap is larger out of domain,
consistent with NGF's prediction that procedural execution is more fragile under length
extension.}
\label{fig:rg-pg-dissociation}
\end{figure}

\begin{figure}[t]
\centering
\includegraphics[width=0.90\linewidth]{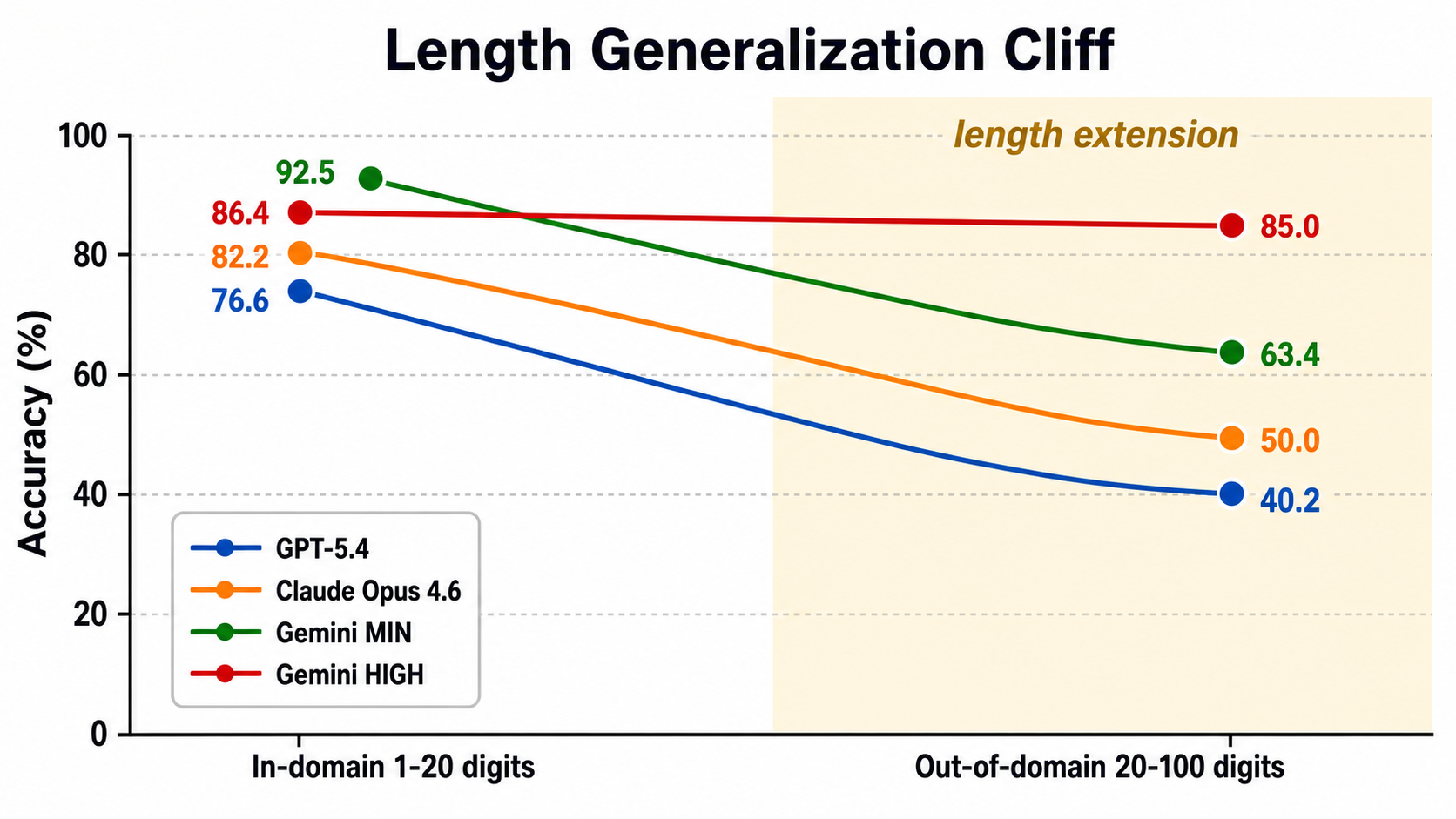}
\caption{\textbf{Length generalization cliff.} Exact-match performance drops sharply when
operand length exceeds the training distribution. Gemini HIGH is the notable exception,
showing that extended reasoning can compensate for some PG failures at substantial inference
cost.}
\label{fig:length-cliff}
\end{figure}

The main dissociation result is robust: every model configuration shows higher RG than PG,
with an average RG--PG gap of approximately 0.19 in domain and 0.27 out of domain. This is not
merely a level shift. Model rankings vary by dimension: Claude is closer to GPT on RG than on
PG, while Gemini MINIMAL leads on atomic primitives but does not transfer that advantage to
all contextual tasks. These results support the central claim that numerical competence is not
monolithic.

\subsection{Reasoning-Budget Tradeoff}
\label{sec5_reasoning}

\begin{table}[t]
\centering
\caption{Gemini~3 accuracy and per-request token usage under MINIMAL vs. HIGH thinking budget.
HIGH improves out-of-domain accuracy but consumes $21.3\times$ as many tokens per request.}
\label{tab:gemini-thinking}
\setlength{\tabcolsep}{5pt}
\begin{tabular}{lcccccc}
\toprule
\textbf{Variant} & \textbf{Exact (in)} & \textbf{Exact (out)} & \textbf{Input} & \textbf{Thinking} & \textbf{Output} & \textbf{Total} \\
\midrule
Gemini MINIMAL & 92.5 & 63.4 & 124 & 0     & 169 & 293 \\
Gemini HIGH    & 86.4 & 85.0 & 125 & 6{,}084 & 38 & 6{,}247 \\
\bottomrule
\end{tabular}
\end{table}

\begin{figure}[t]
\centering
\includegraphics[width=0.95\linewidth]{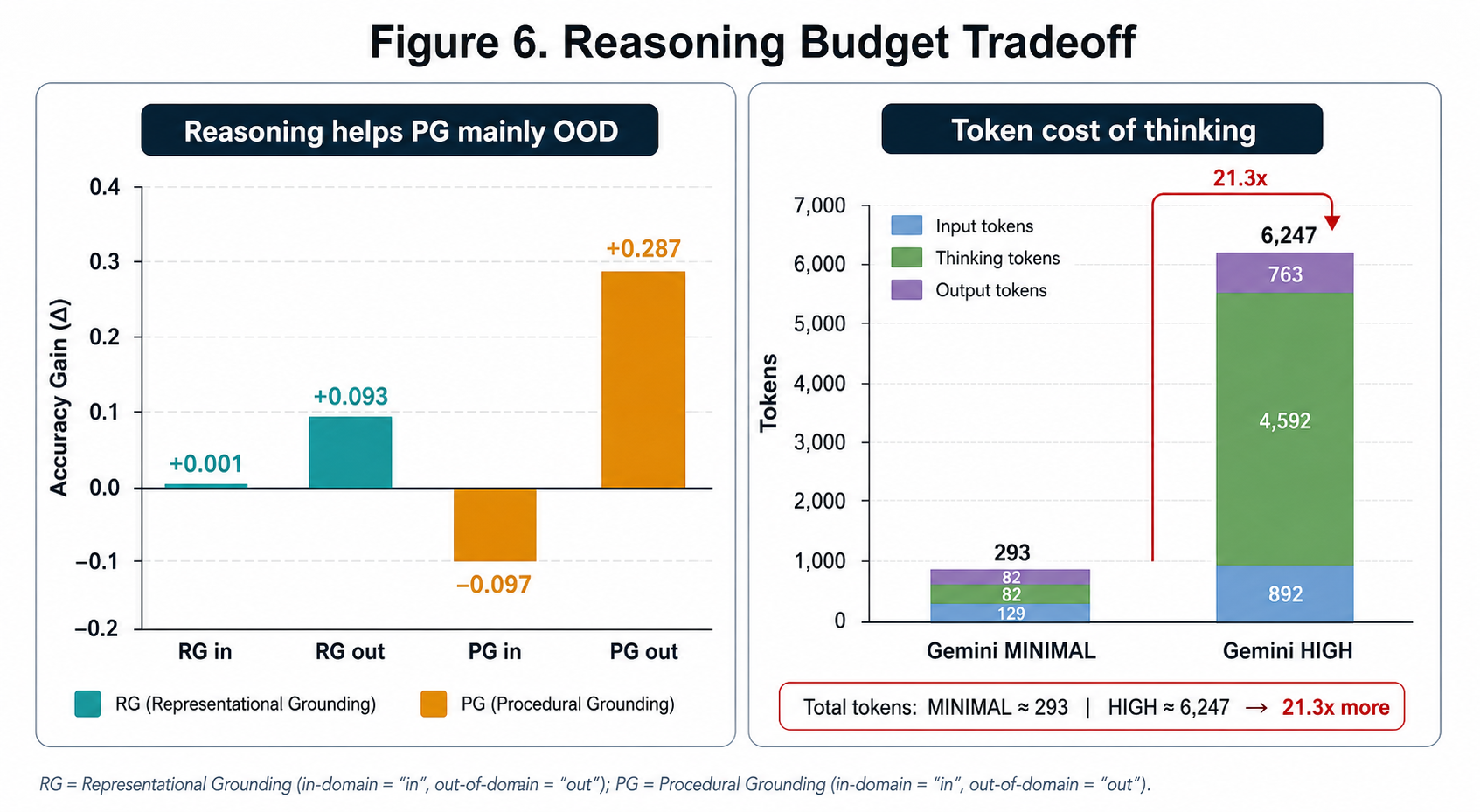}
\caption{\textbf{Reasoning-budget tradeoff.} Extended reasoning disproportionately improves
out-of-domain PG (+0.287) relative to RG (+0.093), while slightly hurting in-domain PG. This
supports the interpretation of reasoning as procedural scaffolding rather than as a structural
fix for tokenization or embedding geometry.}
\label{fig:reasoning-tradeoff}
\end{figure}

The out-of-domain PG improvement is almost three times the RG improvement, exactly the
asymmetry predicted by NGF: Chain-of-Thought supplies external working memory for carry
propagation, column alignment, and intermediate computation, but it cannot alter the input
representation. The in-domain regression is equally important for deployment: on tasks already
inside the well-learned region, extra reasoning can introduce new errors.

\subsection{Tokenizer-Specific RG Blind Spots}
\label{sec5_tokenizer}

\begin{figure}[t]
\centering
\includegraphics[width=0.86\linewidth]{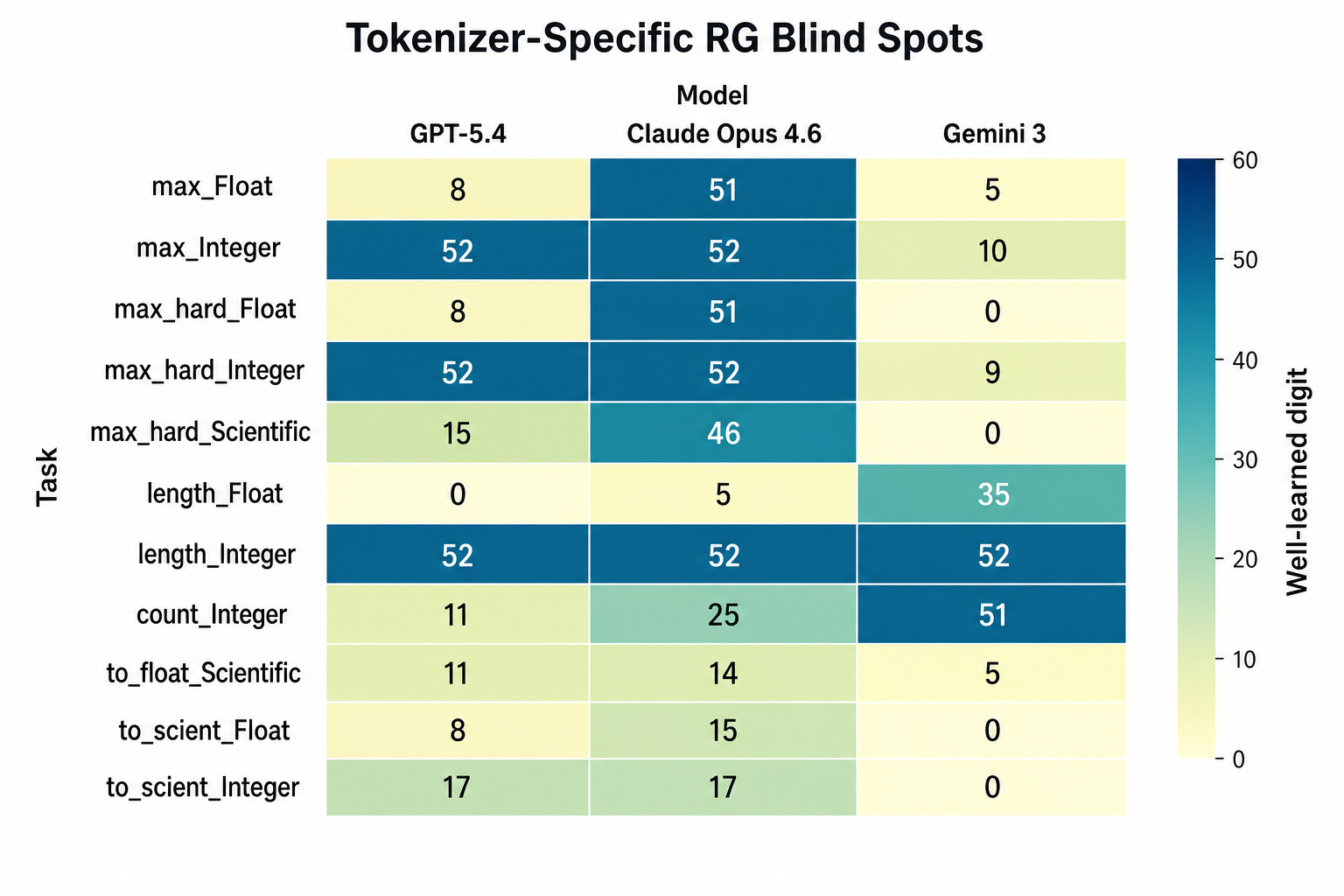}
\caption{\textbf{Tokenizer-specific RG blind spots.} The maximum digit length at which a model
maintains at least 90\% exact-match accuracy varies sharply by task and model. Claude is strong
on float comparison, Gemini is strong on digit counting and length, and GPT remains strong on
integer comparison. These contrasts are consistent with tokenizer-specific surface-form
exposure rather than a single global numeracy score.}
\label{fig:tokenizer-blindspots}
\end{figure}

Tokenizer differences produce qualitatively different RG profiles. For example, Claude Opus~4.6
maintains high accuracy on float comparison to much longer lengths than GPT-5.4 or Gemini~3,
whereas Gemini leads on counting and length tasks. This pattern supports the BPE-bottleneck
hypothesis: different tokenizers create different visibility regimes for place value, decimal
structure, and digit positions.

\subsection{Contextual Numeracy and Fragility}
\label{sec5_context}

\begin{table}[t]
\centering
\caption{Contextual numerical reasoning accuracy on selected NumericBench subsets. The ranking
inverts relative to Number Cookbook: GPT-5.4 leads in contextual arithmetic despite weaker
atomic Number Cookbook performance.}
\label{tab:numericbench}
\setlength{\tabcolsep}{4pt}
\begin{tabular}{lcccc}
\toprule
\textbf{Subset} & \textbf{GPT-5.4} & \textbf{Claude Opus 4.6} & \textbf{Gemini MIN} & \textbf{Gemini HIGH} \\
\midrule
Arithmetic (PG)              & 84.2 & 71.0 & 77.0 & 77.2 \\
Context arithmetic (PG+RG)   & 84.8 & 73.2 & 77.2 & 77.0 \\
Different-digit arithmetic   & 78.8 & 68.5 & 71.2 & 70.6 \\
Num list 100 (RG)            & 98.3 & 33.9 & 66.3 & 69.2 \\
\bottomrule
\end{tabular}
\end{table}

\begin{table}[t]
\centering
\caption{GSM-Symbolic accuracy. Gemini MINIMAL collapses without extended reasoning, while
GPT-5.4 is both accurate and comparatively robust to numerical substitution.}
\label{tab:gsm-symbolic}
\setlength{\tabcolsep}{5pt}
\begin{tabular}{lcccc}
\toprule
\textbf{Split} & \textbf{GPT-5.4} & \textbf{Claude Opus 4.6} & \textbf{Gemini MIN} & \textbf{Gemini HIGH} \\
\midrule
Main           & 96.8 & 75.7 & 26.4 & 78.2 \\
P1 added clause& 94.3 & 85.5 & 10.5 & 56.3 \\
\bottomrule
\end{tabular}
\end{table}

NumericBench and GSM-Symbolic reveal that isolated numerical primitives do not fully predict
numerical reasoning in prose. GPT-5.4 trails on Number Cookbook exact-match accuracy but leads
on NumericBench arithmetic and GSM-Symbolic, suggesting stronger contextual parsing and
instruction following. Gemini MINIMAL shows the opposite pattern: strong atomic performance but
weak word-problem deployment. This motivates an extension of NGF with a third deployment
axis: the ability to activate numerical grounding inside natural-language context and identify
which quantities are relevant.

\subsection{Cross-Benchmark Synthesis}
\label{sec5_synthesis}

\begin{figure}[t]
\centering
\includegraphics[width=0.92\linewidth]{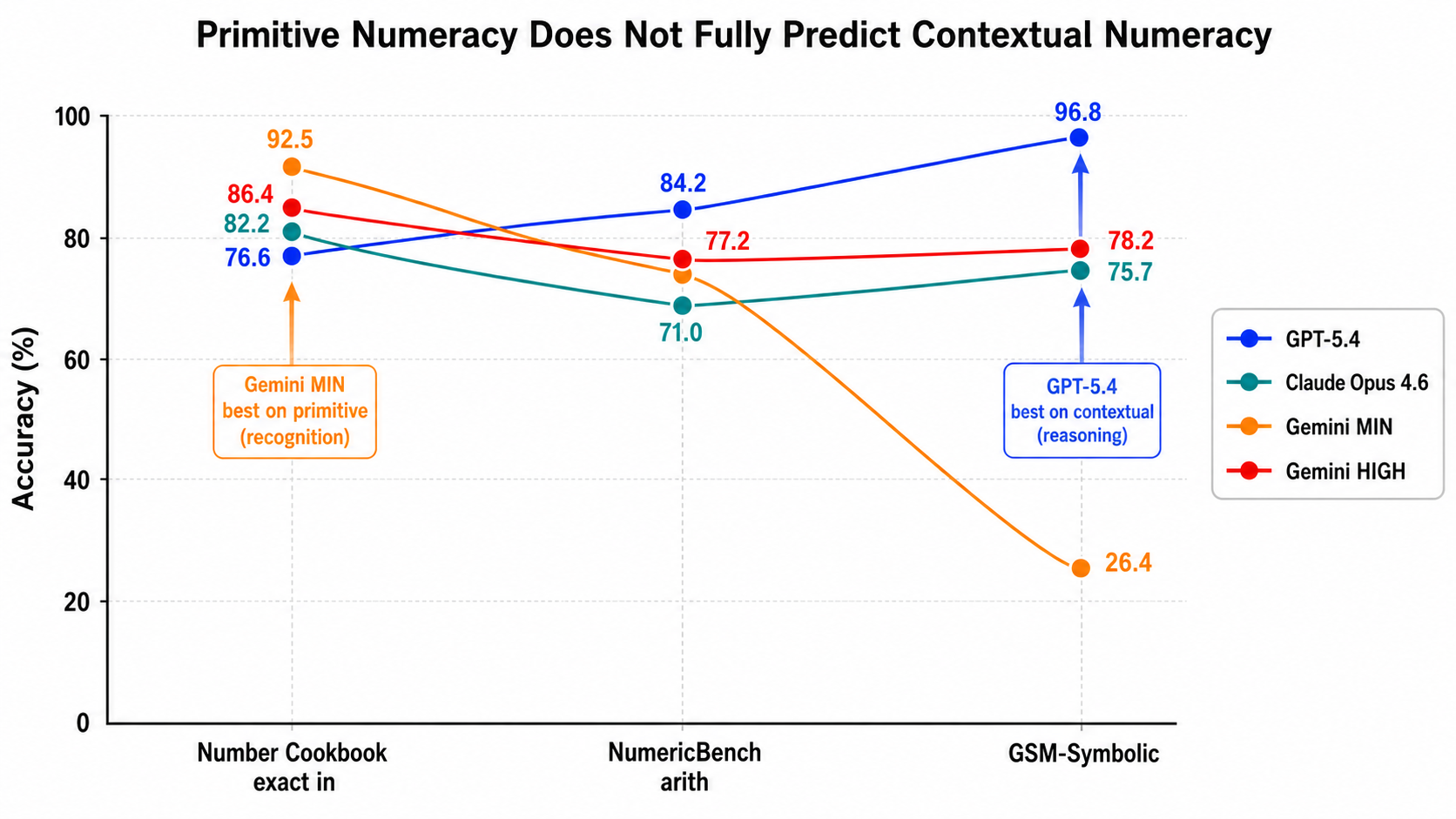}
\caption{\textbf{Primitive numeracy does not fully predict contextual numeracy.} Model ranking
changes across Number Cookbook, NumericBench, and GSM-Symbolic. Atomic RG/PG competence is
necessary but insufficient; contextual deployment mediates whether grounding is actually used
in realistic tasks.}
\label{fig:cross-benchmark-ranking}
\end{figure}

Across the three benchmark families, four conclusions emerge. First, RG and PG are reliably
dissociable. Second, reasoning compensates PG preferentially, but only when the bottleneck is
procedural decomposition rather than surface parsing. Third, tokenizer design shapes RG in a
model-specific manner. Fourth, primitive numerical competence and contextual numerical
competence are only partially coupled. The final point marks the main boundary of the current
NGF formulation and motivates future work on contextual deployment as a third dimension.

\FloatBarrier

\section{Mitigation Strategies and Paths to Improvement}
\label{sec6}

Recognizing the fundamental limitations described above, the research community is pursuing
a multi-layered approach to improving LLM numeracy. We organize the mitigation literature by
the grounding dimension each strategy primarily addresses.

\subsection{The Pretrained-Model Constraint}
\label{sec6_constraint}

Before examining individual strategies, we highlight a critical cross-cutting finding:
\textbf{architectural and tokenization-level interventions that dramatically improve models
trained from scratch are frequently inapplicable or ineffective when applied to large,
already-pretrained models}. We call this the \emph{Pretrained-Model Constraint}.

The evidence is consistent across the key 2024--2025 papers:

\begin{itemize}
\item \textbf{LEFT and Abacus Embeddings}~\citep{mcleish2024transformers}: Both methods require
modifying the model's training regime or architectural components from the start.

\item \textbf{xVal}~\citep{golkar2023xval}: Adopting xVal for a pretrained model requires
replacing the embedding and output head for all number tokens---a cost comparable to
pretraining.

\item \textbf{Digit-level tokenization}: Switching from BPE to character-level or digit-level
tokenization for numbers (as done in Llama~3 \citep{dubey2024llama}) provides measurable RG
benefits but must be done during initial vocabulary construction and pretraining.

\item \textbf{Synthetic fine-tuning}~\citep{wang2024number,tang2024mathscale}: In contrast,
supervised fine-tuning on diverse numerical examples is consistently effective for pretrained
models and represents the most practically accessible intervention.
\end{itemize}

The practical implication is that the most structurally correct solutions are only available
to those training models from scratch, while practitioners working with existing pretrained
models must rely on fine-tuning and inference-time strategies. This creates a two-tier
intervention landscape, which Table~\ref{tab:mitigation} makes explicit.

\subsection{Addressing RG Failures}
\label{sec6_rg}

\subsubsection{Digit-Level and Fixed-Span Tokenization}

The clearest direct intervention is to abandon BPE for numerical tokens entirely.
Character-level tokenization produces a strictly monotone, format-invariant surface
representation. Llama~3 \citep{dubey2024llama} adopts a fixed three-digit tokenization. The
tradeoff is increased sequence length for long numerals, but the RG benefit for magnitude
comparison and format conversion appears robust.

\subsubsection{Continuous Embeddings: xVal}

xVal \citep{golkar2023xval} replaces all number tokens with a single shared \texttt{[NUM]}
embedding, then multiplies it element-wise by the number's scalar value at the embedding
layer. This enforces continuity and is highly effective for scientific data, but poorly suited
for tasks requiring precise digit manipulation.

\subsubsection{Format-Sensitive Fine-Tuning}

For pretrained models where tokenizer replacement is not feasible, targeted fine-tuning on
format-conversion and magnitude-comparison tasks provides a partial remedy, most successful
within the trained format distribution rather than extrapolating across formats.

\subsection{Addressing PG Failures}
\label{sec6_pg}

\subsubsection{Little-Endian Fine-Tuning (LEFT)}

The LEFT strategy \citep{mcleish2024transformers} addresses the right-to-left dependency
problem by training models on \emph{reversed} number strings. In the reversed (Little-Endian)
format, the model generates the least significant digit first, naturally aligning the
generation order with the causal direction of the arithmetic algorithm. LEFT achieves
near-perfect accuracy on addition tasks when applied to models trained from scratch.
Limitations: addition only, scratch-train required, and a post-processing reversal step is
needed for deployment.

\subsubsection{Abacus Embeddings}

Abacus Embeddings \citep{mcleish2024transformers} introduce a specialized positional embedding
for digit tokens that encodes place value relative to the number, injected at every transformer
layer. Combined with a recurrent (looped) transformer architecture, Abacus Embeddings enable
models trained on 20-digit arithmetic to generalize to 100-digit inputs---a $5\times$
improvement in length generalization.

\subsubsection{Curriculum Learning via GSM-Ranges}

The GSM-Ranges dataset systematically varies the magnitude of numbers across several orders
of magnitude, teaching the model that arithmetic is scale-invariant. This reduces but does not
eliminate degradation at larger scales, as the underlying tokenization inconsistency persists.

\subsubsection{Process Reward Models}

\citet{lightman2023lets} and \citet{uesato2022solving} show that process-reward training
yields substantially more reliable arithmetic execution than outcome-only training. Process-reward
training is broadly compatible with pretrained models and is currently the most promising
scalable approach to PG improvement without architectural changes.

\subsubsection{BitTokens}

BitTokens encodes numbers using their IEEE~754 binary floating-point representation. This
teaches the model machine-level arithmetic structure but requires a specialized vocabulary
and output decoding head, shifting the learning burden from decimal arithmetic to bitwise
logic.

\subsection{Inference-Time Strategies (Help Both RG and PG)}
\label{sec6_inference}

\subsubsection{Chain of Thought and Reasoning Models}

The most widely adopted numeracy mitigation is Chain-of-Thought (CoT) prompting
\citep{wei2022chain,nye2021scratchpad}. By generating intermediate tokens before the final
answer, CoT provides external working memory. Reasoning models such as o1
\citep{openai2024o1} and DeepSeek-R1 \citep{deepseek2025r1} are trained via reinforcement
learning to generate extended CoT traces, yielding the single largest performance leap in
the empirical results of Section~\ref{sec5_reasoning}.

The tradeoff is efficiency. Our Gemini thinking-budget measurement
(Table~\ref{tab:gemini-thinking}, Figure~\ref{fig:reasoning-tradeoff}) quantifies the cost:
HIGH thinking consumes $21.3\times$ the
tokens of MINIMAL while regressing on in-domain tasks---a controlled empirical signature of
the ``overthinking'' pathology.

\subsubsection{Tool Use (Program-Aided Language Models)}

For high-reliability numerical applications, the emerging consensus favors Tool Use
\citep{gao2023pal}. The LLM acts as a semantic parser translating the query into executable
code that is then run by a deterministic interpreter. The computed result is exact by
construction---a PG guarantee delivered by externalization rather than by improving the model.
The key limitation is that Tool Use does not eliminate RG requirements; it relocates them.

\subsubsection{Self-Consistency}

Self-consistency \citep{wang2022selfconsistency} draws multiple independent reasoning paths
and aggregates the answers, suppressing single-trace stochastic errors at a cost of additional
inference compute.

\subsection{Comparison of Mitigation Strategies}
\label{sec6_comparison}

Table~\ref{tab:mitigation} provides a structured comparison of the mitigation strategies
discussed above, organized by the three dimensions most relevant to practitioners:
pretrained-model compatibility, grounding dimension primarily addressed, and primary
limitation.

\begin{figure}[t]
\centering
\includegraphics[width=0.92\linewidth]{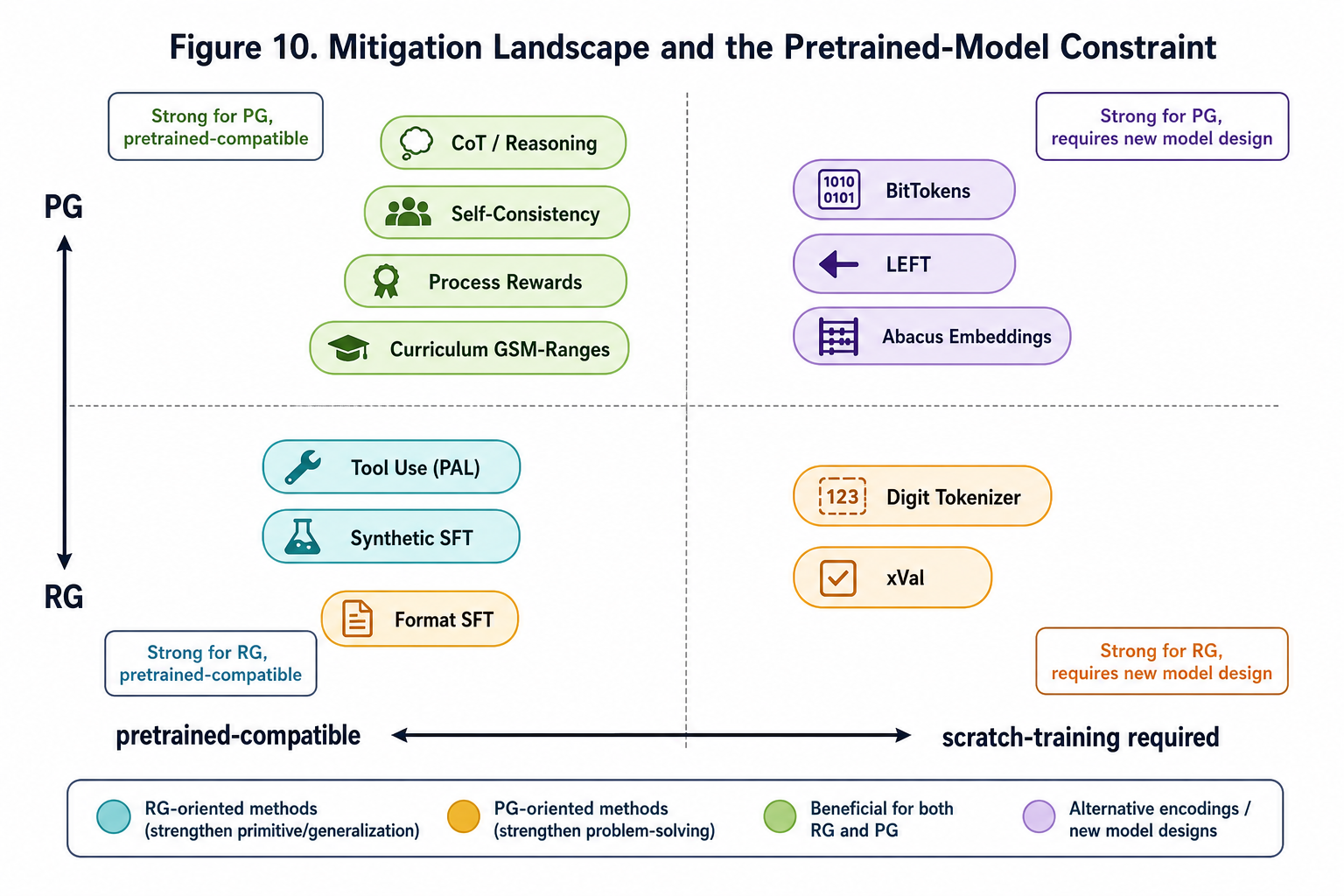}
\caption{\textbf{Mitigation landscape and the Pretrained-Model Constraint.} The most structural
fixes for RG and PG lie on the scratch-training side of the matrix. Practitioners working with
existing pretrained models mostly operate on the left side: supervised fine-tuning, process
rewards, reasoning, self-consistency, and tool use.}
\label{fig:mitigation-landscape}
\end{figure}

\begin{table}[htbp]
\centering
\caption{Comparison of NGF mitigation strategies. \textbf{Pretrained Compat.}: whether the
method can be applied to an existing pretrained model without retraining from scratch.
\textbf{Grounding}: primary grounding dimension addressed (RG, PG, or Both). SFT~=~Supervised
Fine-Tuning.}
\label{tab:mitigation}
\setlength{\tabcolsep}{3pt}
\renewcommand{\arraystretch}{1.08}
\footnotesize
\begin{tabularx}{\textwidth}{@{}
  >{\raggedright\arraybackslash}p{2.5cm}
  >{\centering\arraybackslash}p{1.6cm}
  >{\centering\arraybackslash}p{1.8cm}
  >{\centering\arraybackslash}p{1.8cm}
  >{\raggedright\arraybackslash}X@{}}
\toprule
\textbf{Method} & \textbf{Category} & \shortstack{\textbf{Pretrained}\\\textbf{Compat.}} &
\textbf{Grounding} & \shortstack{\textbf{Primary}\\\textbf{Limitation}} \\
\midrule
Digit / fixed-span tokenization & Arch. & No  & RG  & Scratch-train required; longer sequences \\
xVal                            & Arch. & No  & RG  & Not for exact digit-level tasks \\
Format-sensitive SFT            & Data  & Yes & RG  & Generalizes weakly across formats \\
\midrule
LEFT                            & Arch. & No  & PG  & Addition only; scratch-train required \\
Abacus Embeddings               & Arch. & No  & PG  & Scratch-train required; $+5\times$ length gen. \\
Curriculum (GSM-Ranges)         & Data  & Yes & PG  & Doesn't fix tokenization-induced RG \\
Process Reward Models           & Train.& Yes & PG  & Requires step-labeled training data \\
BitTokens                       & Arch. & No  & PG  & New vocabulary and decoding head required \\
\midrule
SFT on Synthetic Data           & Data  & Yes & Both & Does not fix tokenization artifacts \\
CoT / Reasoning (o1, R1)        & Inf.  & Yes & Both & ${\sim}18\times$ token cost; overthinking \\
Tool Use (PAL)                  & Inf.  & Yes & Both & Relies on correct equation formulation (RG) \\
Self-Consistency                & Inf.  & Yes & PG   & Additional inference cost \\
\bottomrule
\end{tabularx}
\end{table}

\subsection{Practical Deployment Recommendations}
\label{sec6_practical}

\begin{figure}[t]
\centering
\includegraphics[width=0.88\linewidth]{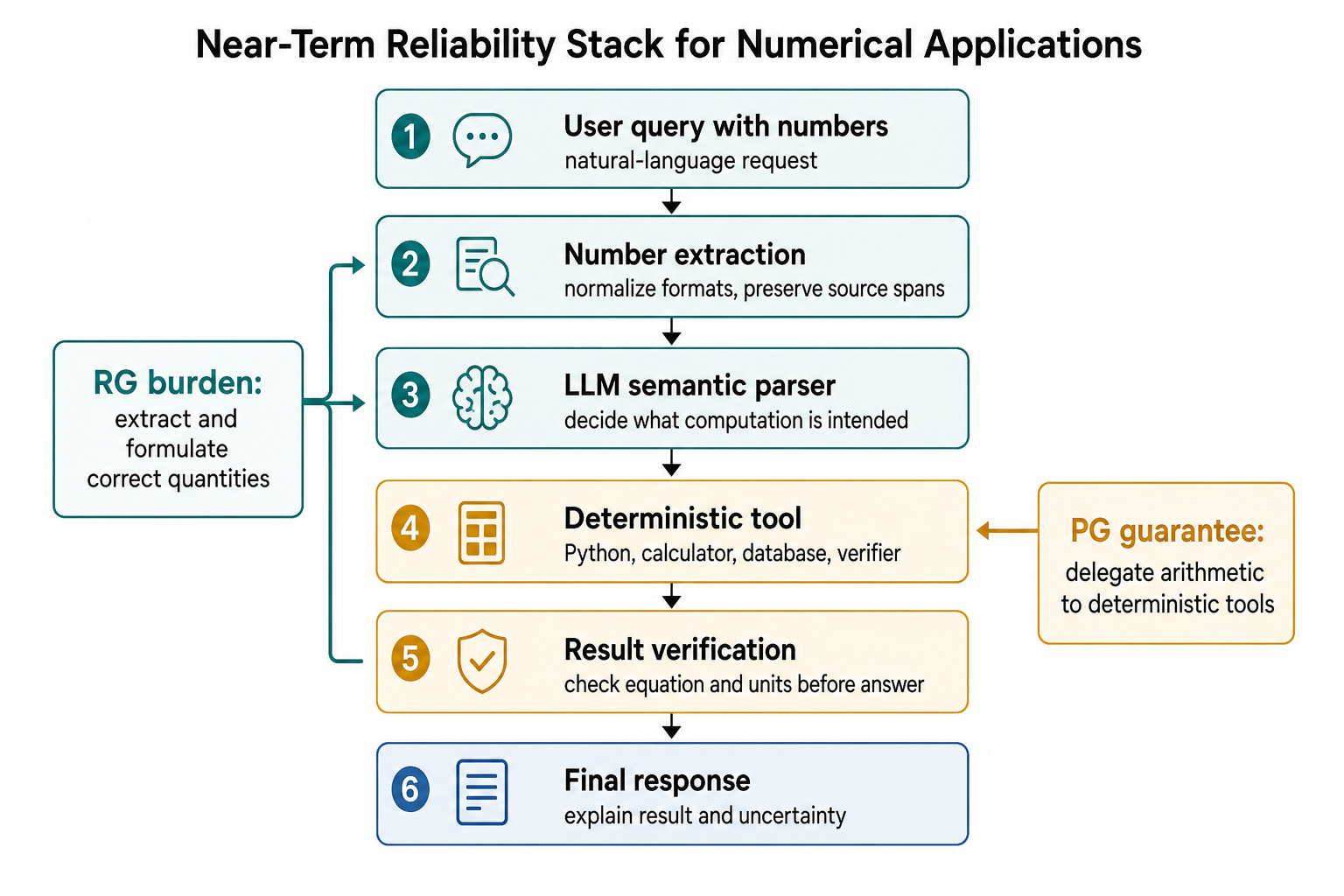}
\caption{\textbf{Near-term reliability stack for numerical applications.} Tool use solves PG
by delegating calculation to deterministic machinery, but it does not remove RG: the model
must still extract quantities, preserve formats and units, and formulate the correct equation.}
\label{fig:deployment-stack}
\end{figure}

\textbf{High-reliability numerical computation (finance, engineering, medicine).}
Default to Tool Use. Instruct the model to generate code rather than numeric answers directly,
and execute the code in a sandboxed interpreter. Verify that the generated equation is correct
before trusting the output.

\textbf{Mixed reasoning and computation tasks.}
Use CoT prompting with a reasoning model. Insert an explicit verification step or tool call
for any intermediate numerical result that drives subsequent reasoning.

\textbf{Numerical understanding in natural language.}
RG tasks are better supported in reasoning models than in generalist models. For the highest
reliability, extract all numbers as a preprocessing step and pass them as a structured list.

\textbf{Fine-tuning for domain-specific numeracy.}
Supervised fine-tuning on domain-relevant synthetic examples provides reliable improvement
within the trained range. Do not assume that fine-tuning generalizes beyond the numerical
range represented in the fine-tuning data.

\textbf{Benchmark scores as a floor, not a ceiling.}
Always evaluate grounding directly on data representative of the target deployment context
before drawing conclusions from aggregate benchmark scores.

\section{Conclusions and Future Directions}
\label{sec7}

The investigation into numeracy in Large Language Models reveals a field in transition. The
gains available from naive scaling are insufficient to fix the tokenization blindness or
compensate for the absence of arithmetic working memory. The $9.11 > 9.9$ error is not a
random glitch; it is a structural artifact of processing numbers as linguistic tokens rather
than as values on an ordered continuum---a failure of \emph{representational grounding} in
the precise sense of NGF.

\subsection{Key Takeaways}

\textbf{Numerical grounding is architecturally distinct from mathematical reasoning.}
High scores on mathematical reasoning benchmarks measure a model's ability to structure and
follow a logical argument, not its ability to reliably execute the arithmetic primitives on
which that argument depends. NGF predicts the direction of dissociation: reasoning training
improves PG more than RG.

\textbf{Tokenization is the primary structural bottleneck for RG.}
BPE is fundamentally ill-suited for arithmetic. The most durable progress on RG requires
digit-level tokenization, place-value-aware embeddings (Abacus), or number-specific
representation schemes (xVal, BitTokens)---all integrated at the pretraining stage.

\textbf{Reasoning is a powerful compensatory mechanism for PG, not a structural fix for RG.}
CoT and RL-trained reasoning models dramatically improve numerical performance but are
inefficient and degrade on trivial computations. Our Gemini thinking-budget measurement
provides a controlled
empirical demonstration. CoT cannot fix surface-to-value mappings already mis-tokenized at the
input layer.

\textbf{Tool Use is the near-term reliability solution for PG.}
For any application where numerical correctness is safety-critical, delegate arithmetic to a
deterministic external tool. The LLM's role is semantic parsing (an RG task), not
calculation.

\textbf{The Pretrained-Model Constraint defines the practical landscape.}
The most structurally correct solutions are available only to those training new models from
scratch. For the broader community, viable interventions are limited to fine-tuning,
process-reward training, and inference-time strategies.

\subsection{Future Directions}

\textbf{Algorithmic generalization across arbitrary lengths.}
Hybrid approaches that combine Abacus-style inductive biases with large-scale pretraining
represent the most promising direction for PG, but integration at scale remains open.

\textbf{Unified number representation.}
A ``number-aware tokenizer'' that segments numbers consistently into digits or
representation-aware sub-units would eliminate the structural root cause of most RG failures.

\textbf{Cross-lingual and cross-format numerical reasoning.}
Current diagnostic benchmarks are almost entirely English and decimal; MGSM
\citep{shi2022language} is the principal exception, but covers only word problems.

\textbf{Process-level reward for arithmetic.}
Process reward models \citep{lightman2023lets,uesato2022solving} provide a more targeted
training signal for building reliable PG without the overthinking pathology of current
reasoning models, and integrate cleanly into existing pretrained-model fine-tuning pipelines.

\textbf{Bridging cognitive-science and machine grounding.}
NGF's RG/PG decomposition mirrors human dual-system numerical cognition
\citep{dehaene2011number,spelke2007core}. Future work that operationalizes the
cognitive-science distinctions more precisely---approximate vs.\ exact, log-scaled vs.\
linear-scaled, perceptual vs.\ symbolic---could yield richer grounding tests and more targeted
architectural prescriptions.

The path toward robust, human-like number sense in foundation models requires coordinated
progress on tokenization, positional encoding, training objectives, and pretraining data
composition. Until these structural issues are resolved---at both the representational and
procedural levels of grounding---LLMs will remain powerful reasoning scaffolds but unreliable
calculators.

\section*{AI Assistance Disclosure}
AI-assisted tools were used during manuscript preparation for literature-search support,
structure planning, language editing, figure design, review, and formatting. All content,
arguments, and conclusions were directed and reviewed by the author, who takes full
responsibility for the accuracy and integrity of this work.

\bibliography{main}
\bibliographystyle{tmlr}

\end{document}